\documentclass[twoside,11pt]{article}
\usepackage{jair, theapa, rawfonts}
\usepackage{graphicx}
\usepackage{caption}
\usepackage{subfigure}
\usepackage{amsmath, amssymb}
\usepackage{makecell}
\usepackage{multirow}
\ShortHeadings{GP: Context-free Grammar Pre-training for Text-to-SQL Parsers}
{Zhao Liang, Cao Hexin, \& Zhao Yunsong}
\firstpageno{1}

\begin{document}

\title{GP: Context-free Grammar Pre-training for Text-to-SQL Parsers}

\author{\name Zhao Liang \email zhaoliang146@ocft.com \\
       \name Cao Hexin \email caohexin771@ocft.com \\
       \name Zhao Yunsong \email zhaoyunsong244@ocft.com \\
       \addr OneConnect Financial Technology Big Data Lab\\
       Shanghai, China
       }


\maketitle

\begin{abstract}
Text-to-SQL technology has significant applications in realizing database query through natural language, with no requirement for learning SQL grammar. Nevertheless, the challenge is that modeling alignment between database information and consideration in a certain query is not obvious. Text-to-SQL parsing is proposed as novel Grammar Pre-training (GP) to decode deep relations between database and question. To adequately learn the internal relationship of SQL grammar, the decoder is pre-trained independently of the encoder. Subsequently, the robustness of the model is improved and convergence is accelerated. Flooding level is adopted to obtain the non-zero training loss and avoid local extrema problems. Ultimately, we achieved better performance on Spider, a cross-DB Text-to-SQL dataset (72.8\% dev, 69.8\% test)by encoding the sentence with ${\bf G{\scriptstyle RAPPA}}$ and RAT-SQL model. By reducing the average loss by 78.9\%, the variance is only 0.8\% of the previous model while training. Moreover, experiments proved that this technique converges much faster and has excellent robustness.
\end{abstract}


\section{Introduction}
\label{Introduction}

Recently, with the development of artificial intelligence technology, to directly generate SQL statements has attracted a huge deal of research interest. These statements interact with database systems through the analysis of natural language. A Natural Language Interface to Database (NLIDB)is adopted by current research work to realize the interaction between user's questions and the database system to obtain and analyze data\cite{baik2019bridging}.

The core problem of NLIDB is to convert the input text information into SQL statements (Text-to-SQL). For solving this problem, two main approaches exist at present. First, the method is based on a rule template, indicating that the natural language is classified based on the common SQL grammar. Therefore, the corresponding SQL templates is related to various categories\cite{popescu-etal-2004-modern,unger2012template}. Such a method requires manual summarization of experience and a huge deal of time\cite{li2014constructing}. Moreover, by changing the application scenario, the existing templates are often difficult to satisfy the requirements. Hence, the migration is poor. Second, based on the deep learning method, the neural network is utilized for end-to-end implementation\cite{zhong2017seq2sql,yu2018typesql,yu2018syntaxsqlnet,bogin2019global,guo2019towards}. This approach can be self-optimized by continuously adding the sample information. It includes the advantages of both higher accuracy and strong stability and receives further attention from the academic community. By incorporating it with the BERT encoder, the WikiSQL dataset accuracy of above 90\% can be obtained. 
\begin{figure}[htbp]
  \centering
  \includegraphics[width=0.8\textwidth]{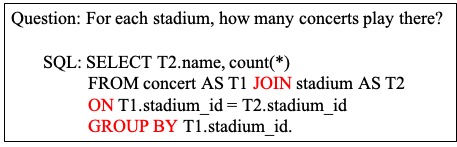}
  \caption{Complex SQL statement for multi-table connection in Spider dataset}
  \label{fig:1st}
\end{figure}

However, satisfactory performance is not achieved by these deep-learning methods on a cross-domain Text-to-SQL scenario such as Spider\cite{yu2018spider}. According to Fig. \ref{fig:1st}, this SQL query includes nested clauses such as GROUP BY and connections in multiple tables. Such grammar details are concerned rarely by users, hence, they are hardly mentioned in questions. Bailin Wang et al. proposed a relation-aware framework called RAT-SQL and achieved state-of-art accuracy on the Spider dataset. Moreover, pre-training language models are developed based on structured table data and the natural language of users. At early stages, BERT\cite{devlin2019bert} and RoBERTa\cite{liu2019roberta} for contextual sentences are used in cross-domain Text-to-SQL scenario, however, the relation between the tables and fields of the database is not considered. A grammar-augmented pre-training model (${\bf G{\scriptstyle RAPPA}}$) is presented describing the joint representations of textual and tabular data \cite{yu2020GRAPPA}. By integrating the pre-training model with other downstream methods such as RAT-SQL, the accuracy of cross-domain tasks can be improved greatly.


In the present work, a context-free grammar pre-training (GP) method is proposed creatively for Text-to-SQL. Since SQL grammar framework is irrelevant to the specific natural language, we first pre-trained the decoder without encoder information. Within the training step, GP effectively improves the training efficiency of the model and has good advantages in robustness and convergence. Within the preprocessing module, we used string matching to discover the value appearing in the question, and add it behind the equivalent column on the original input sequence. To design the loss function, we adopted flooding level as a new method to avoid local minimum values. Based on ${\bf G{\scriptstyle RAPPA}}$/RAT-SQL framework, experiments indicated that a much higher accuracy on Spider dataset and better robustness is obtained by our approach. It also presents potential applications for other context-free grammar representation tasks.

\section{Related Works}

{\bf Pre-training models for NLP parsing}  Text-to-SQL task comprise both structured schema information and unstructured user question. Early research used general pre-training models such as Elmo\cite{peters2018deep}, RoBERTa\cite{liu2019roberta}, and BERT\cite{devlin2019bert} to represent textual information for unstructured language questions. There has been a great enhancement in the joint textual-tabular field like question answering\cite{chen2020hybridqa} and table semantic parsing\cite{yu2018spider} by learning better representations from the input text and table information. However, they mostly consider single tables. In recent pre-training work, it is focused on achieving high-quality cross-modal representation. TaBERT \cite{yin2020tabert} is pre-trained through millions of web tables. It can denote complete structure for various tables and make some matrix computations in table semantic parsing. Nevertheless, its performance is weakened by the noisy context information on the Text-to-SQL task. In this work, we adopt ${\bf G{\scriptstyle RAPPA}}$, the grammar-augmented pre-training technique utilizing a novel text-schema link objective and masked language modeling (MLM). Integrating ${\bf G{\scriptstyle RAPPA}}$ as feature representation layers with other downstream models, great accuracy is obtained on the Spider dataset. \\
{\bf Neural networks for Text-to-SQL}  Previous networks are intended to solve problems in single table dataset such as WikiSQL. The Seq2SQL model based on the strategy mode\cite{zhong2017seq2sql} is used in Text-to-SQL tasks and SQL execution accuracy of 59.45\% is achieved on the WikiSQL dataset. Then, TypeSQL\cite{yu2018typesql} is presented to further extract the keywords in the question sentence by integrating external knowledge and database field enumeration values. The obvious results were obtained by the above method in a single table query, however, it is not enough for solving the complex mode of the multi-table query. EditSQL\cite{zhang2020editing} utilizes an editing mechanism to introduce historical information for user queries, moreover, its matching accuracy on Spider dataset reaches up to 32.9\%. an intermediate representation called SemQL is used in IRNet\cite{guo2019towards} to translate complex SQL queries into a syntax tree. Using pointer network\cite{vinyals2015pointer} for downstream tasks, an accuracy of 54.7 is obtained on the Spider test set. Moreover, graph neural networks are concerned to present the relations for schema information. Global gated graph neural network\cite{bogin2019global} is designed to train the database patterns’ structure and apply it in the encoding and decoding stages. Recently, RAT-SQL \cite{wang2020rat} used a relation-aware self-attention mechanism for schema encoding, schema linking, and feature representation. It obtains the state-of-art accuracy of 65.6\% on the Spider test set. \\
{\bf Training loss optimization}  $Overfitting$ is a common problem in training procedure \cite{Goodfellow-et-al-2016}. Comparing with former methods such as dropout \cite{srivastava2014dropout}, label smoothing\cite{szegedy2016rethinking} batch normalization \cite{ioffe2015batch}, and mixup\cite{zhang2017mixup}, to avoid the training loss from decreasing to zero, flooding level\cite{ishida2020we} makes the training loss float around a small constant value. On the other hand, the loss fixed around a certain level can be determined based on the model itself. Thus, flooding skips some local extreme points to find the optimal parameters from a global perspective.

\section{Methodology}
\label{mv-heuristic}
\subsection{Context-free Grammar Pre-training}
RAT-SQL uses the {\bf Syntactic Neural Model (SNM)} presented by \cite{yin2017syntactic} to create the SQL grammar. Yin et al. believed that the present methods treat code generation as a task of sequence generation not considering the grammar of the target programming language. Programming languages, especially SQL, have strict grammar rules, unlike natural languages. Based on these rules, {\bf SNM} is essentially a method to improve the accuracy of the model by limiting the search space of the decoder.

Moreover, the basic framework of SQL grammar is context-free with the specific natural language description. For instance, regardless of the natural language description, the first clause of SQL is always $select$, and the next clause is always $from$. Based on the experiments, the loss value in the initial training stage of RAT-SQL is extremely large mainly coming from SQL grammar errors created by the decoder.

Regarding the above situation, we proposed a context-free Grammar Pre-training ({\bf GP}) technique to pre-train the parameters on the decoder side. The encoder’s semantic information is replaced by zero vectors. The probability equation of RAT-SQL utilizing LSTM to output a sequence of $decoder$ actions is:
\begin{equation}
    Pr(P|y) = \prod_{t}{}Pr(a_t|a_{<t},y)
    \label{raw_action}
\end{equation}
where $y$ is always [$\bf{0}$] in the stage of GP and $a_{<t}$ are all previous actions. Correspondingly, the LSTM's state updating strategy will be modified as:
\begin{equation}
    m_t,h_t=f_{LSTM}([\bf a_{t-1} || [0] || h_{p_t} || a_{p_t} || n_{f_t}], m_{t-1}, h_{t-1})
    \label{lstm}
\end{equation}
where $m_{t}$ and $h_{t}$ are the LSTM cell state and output in step $t$, $a_{t-1}$ represents the embedding of the previous action, $p_t$ denotes the step equivalent to expanding the parent AST node of the current node, and $n_{f_t}$ is the current node type embedding. We used $\bf{[0]}$ to replace the former $z_t$ obtained through multi-head attention on $h_{t-1}$ over $y$.

Since GP no longer depends on semantic information, it cannot predict column’s or table’s names. It is assumed that each sample has only one column and one table in order to not change the framework of RAT-SQL, , thus
\begin{equation}
    Pr(a_t=S{\scriptstyle ELECT}C{\scriptstyle OLUMN}[0]|a_{<t})=1
    \label{col}
\end{equation}
\begin{equation}
    Pr(a_t=S{\scriptstyle ELECT}T{\scriptstyle ABLE}[0]|a_{<t})=1
    \label{tab}
\end{equation}
To prevent overfitting, the number of decoder Grammar Pre-training steps was limited to 300. 

\subsection{Question-Schema Serialization and Encoding}
Generally, the serialization technique of RAT-SQL is adopted. Since the utilized pre-trained semantic model is ${\bf G{\scriptstyle RAPPA}}$, the question tokens are preceded by $\langle s \rangle$ and ended up with $\langle/s\rangle$. Then, tables and columns are spliced in sequence based on the order of the schema presented by the Spider dataset. Moreover, we used $\langle/s\rangle$ as the separator.

\begin{figure}[h]
  \centering
  \includegraphics[width=0.8\textwidth]{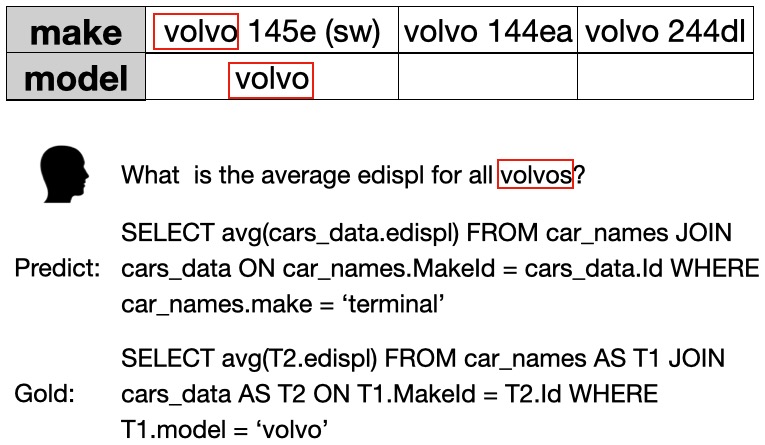}
  \caption{A example from Spider dataset. $volvo$ is denoted in both $make$ and $model$}
  \label{fig:example}
\end{figure}

As stated in \cite{lin2020bridging}, modeling with only table/field names and their relations is not always adequate for capturing the semantics of the schema and its dependencies with the question. Remarkably, we append values to mention columns only when they match the question exactly. For example, in Figure \ref{fig:example}, the keyword $volvo$ in the question appears in both column $make$ and column $model$, respectively. Thus, there is a relationship between the token $volvo$ and a Column-Part-Match(CPM) with column $make$ as well as a Column-Exact-Match(CEM) relationship with column $model$. Intuitively, the exact match possesses a greater probability as the correct column. To strengthen this relationship, we put $volvo$ after the column $model$ during serializing while column $make$ not. The sequence can be converted as
\begin{equation}
S = \left \langle s \right \rangle, Q, \left \langle /s \right \rangle, C_1, \left \langle /s \right \rangle, C_2, V_2, \left \langle /s \right \rangle, ..., T_1, \left \langle /s \right \rangle, T_2, \left \langle /s \right \rangle, ..., \left \langle /s \right \rangle
\end{equation}

\begin{figure*}[h]
  \centering
  \includegraphics[width=\linewidth]{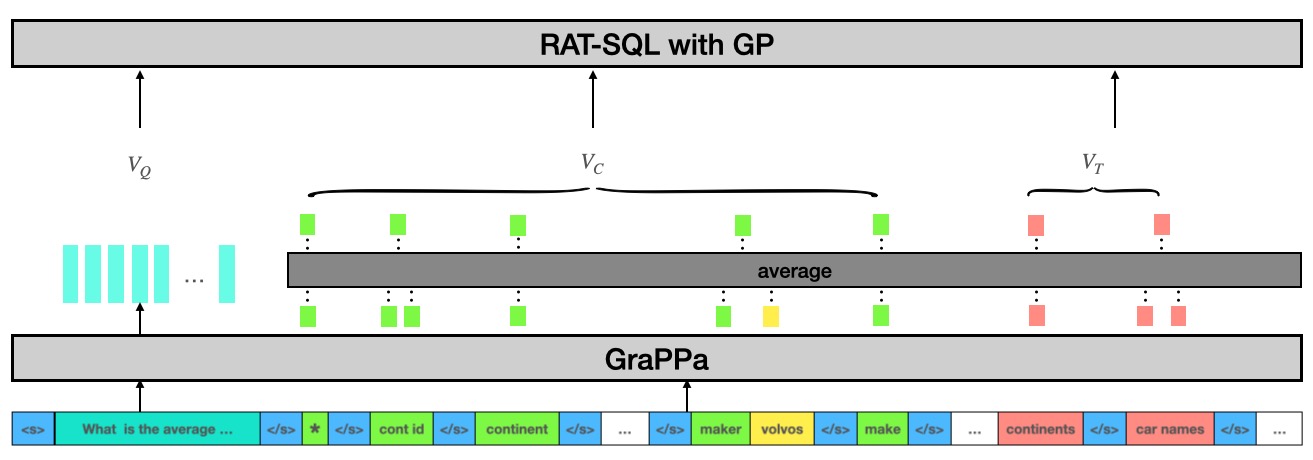}
  \caption{An illustration of encoder model}
  \label{fig:frame}
\end{figure*}

In RAT-SQL, the vector representation of a table or a column is the average of the last and first token. Research indicates that this encoding method may lose important information\cite{wang2020rat}, hence, another technique is utilized by calculating the average of all tokens' vector of the column or table. When a column is followed by a value, the column’s representation is determined by all column tokens and value tokens (Fig.\ref{fig:frame}).

For deep learning, training loss keeps often decreasing while the validation loss suddenly starts to rise\cite{Goodfellow-et-al-2016}. \cite{ishida2020we} proposed a tricky and simple loss function $flooding$ to decrease validation loss continuously:

\begin{equation}
    \tilde{J}(\theta)=|J(\theta)-b|+b
    \label{equation_flooding}
\end{equation}

where $b>0$ represents the user-specified flooding level, and $\theta$ is the model parameter. It is assumed that the existence of parameter $b$ can prevent the model from falling into the local optimum to a certain extent, during the optimization process. Since spider dataset has various types of SQL grammar and databases sizes are usually inconsistent, it usually leads to overfitting and converges near a local extreme while training, here, this method was adopted to make final results well. Nevertheless, unsuitable $b$ usually result in gradient explosion.

\section{Experimental Results}

\subsection{Experimental Setup}
The Adam optimizer\cite{kingma2015adam} with default hyperparameters is adopted. In the stage of GP, the learning rate is set as $7.44\times10^{-4}$. Owing to GPU memory limitation, we set $bs=3$ and $num\_batch\_accumulated=4$, in which, $bs$ and $num\_batch\_accumulated$ are the gradient accumulation parameters of RAT-SQL equivalent to batch size of 12. Considering GP and a smaller batch size, compared to RAT-SQL, we set the initial learning rate of ${\bf G{\scriptstyle RAPPA}}$ from the original $3\times10^{-6}$ to $2\times10^{-6}$, and the initial learning rate of other model parameters from $7.44\times10^{-4}$ to $5.44\times10^{-4}$. The rest of the setups are the same with RAT-SQL.

\subsection{Dataset and Metrics}

\begin{table}[h!]%
  \centering
  \begin{tabular}{ccl}
    \hline 
    dataset & samples & databases\\
    \hline 
    train set & 8659 & 146\\
    dev set & 1034 & 20\\
    test set & 2147 & 40\\
    \hline 
\end{tabular}
\caption{Size of Spider}
\label{tab:spider}
\end{table}

$\bf{Spider}$ \cite{yu2018spider} is a cross-domain Text-to-SQL dataset and large-scale complex. It includes both schema information and a corresponding SQL statement for each natural language problem. According to Table \ref{tab:spider}, it includes 10,181 questions and 5,693 unique complex SQL queries on 206 databases with multiple tables covering 138 different domains. Based on its hardness level, spider splits into 4 types of data sets as Easy, Medium, Hard, and Extra Hard. It is the only data set in the public data set of Text-to-SQL tasks containing both complex SQL statements and multi-table query. Here, the complex SQL denotes the nested query situation of $order by$, $group by$, and $where$ clauses in the statement.

The metric adopted to assess model performance is $\bf{Exact}$ $\bf{Match}$ $\bf{Accuracy}$ suggested by \cite{yu2018syntaxsqlnet}. This metric refers to utilize standardized definitions to process the prediction SQL and the true statement, and calculate matching degree between them, without considering the column names order.

\subsection{Results}
While RAT-SQL and ${\bf G{\scriptstyle RAPPA}}$ are open-sourced, the offline result is worse compared to announced on the leaderboard in our experiments (Table \ref{tab:compare}). The reason can be explained by random seed or device differences. In this section, we mainly compared model performance based on offline results.

\begin{table}[h!]%
  \centering
  \begin{tabular}{ccl}
    \hline 
    model & leaderboard & offline\\
    \hline 
    RAT-SQL+Bert & 69.7 & 66.7\\
    RAT-SQL+${\bf G{\scriptstyle RAPPA}}$ & 73.4 & 71.7\\
    \hline 
\end{tabular}
\caption{A comparison between offline and leaderboard results on dev set}
\label{tab:compare}
\end{table}

\begin{figure}[h]
\centering
\begin{minipage}[h]{0.45\textwidth}
\centering
\includegraphics[width=\textwidth]{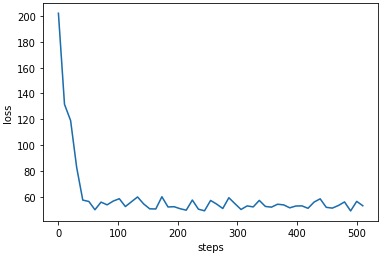}
\caption{Grammar Pre-training loss value curve}
\label{gp_loss}
\end{minipage}
\begin{minipage}[h]{0.45\textwidth}
\centering
\includegraphics[width=\textwidth]{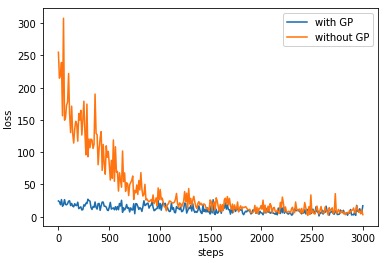}
\caption{Loss curve comparison between with GP and without GP}
\label{comp_loss}
\end{minipage}
\end{figure}

{\bf GP}  Figure \ref{gp_loss} indicates that in the first 50 steps of GP, the training loss significantly drops, then, it remains at about 53. To prevent overfitting, the number of Grammar Pre-training steps is limited, even if the loss is still dropping at a tiny speed. Then, we used the pre-trained decoder to train our model, and the training loss is maintained at a lower level compared to the previous method without GP (Fig. \ref{comp_loss}). We computed the average and variance of loss before and after 1500 steps as stable values. From Table \ref{tab:comparison for loss}, the average loss with GP is 15.02, which is reduced by 78.9\% compared to the former one. Furthermore, its variation rate is only 0.8\% of the model without GP indicating that there is a smooth optimization during training. The final loss of less than 1.37 also proves that GP helps to find auxiliary information between SQL grammar and question words.

\begin{table}[h!]
  \centering
  \begin{tabular}{crrrr}
    \hline
    \multirow{2}*{Model} &
    \multicolumn{2}{c}{$0-1500$} & 
    \multicolumn{2}{c}{$1500-81000$}\\
    \cline{2-5} 
    & Avg & Var & Avg & Var\\ 
    \hline
    Without GP & $71.02$ & $3660.93$ & $1.37$ & $7.35$\\ 
    With GP & $15.02$ & $29.67$ & $1.10$ & $4.47$\\ 
    \hline
\end{tabular}
\caption{Comparison for the average and variance of loss before and after $1500$ steps}
\label{tab:comparison for loss}
\end{table}

{\bf Flooding}  Equation \ref{equation_flooding} indicates that there is an extra parameter $b$ in loss function, and the model performance is extremely sensitive to $b$ and learning rate $lr$. Moreover, a slightly larger $b$ may lead to the model to gradient explosion during training. Table \ref{tab:flooding} indicates several examples about different parameter combination. $\emptyset$ denotes that the parameter combination will result in gradient explosion. It is worth mentioning that although $Flooding$ can enhance model performance, the results are not stable, in which, the best result may be as high as 72.1\%, and the lowest result may be only 70.7\% even if we used the same parameters.

\begin{table}[h]%
  \centering
  \begin{tabular}{cccc}
    \hline 
    $b$ & $lr$ & $bert\_lr$ & Dev.\\
    \hline 
    $0.1$ & $7.44\times10^{-4}$ & $3\times10^{-6}$ & $\emptyset$\\
    $0.2$ & $5.44\times10^{-4}$ & $2\times10^{-6}$ & $\emptyset$\\
    $0.02$ & $5.44\times10^{-4}$ & $2\times10^{-6}$ & $70.6\pm0.6$\\
    $0.01$ & $5.44\times10^{-4}$ & $2\times10^{-6}$ & $71.4\pm0.7$\\
    \hline 
 \end{tabular}
 \caption{The influence of different parameters $b$ and $lr$ on the results. $\emptyset$ indicates that the combination of this parameters will lead to the explosion of the gradient.}
 \label{tab:flooding}
\end{table}

{\bf Serialization with value} By adding the equivalent value after the column, the recognition between columns is enhanced. It is indicated that a slight reduction exists in column selection errors.
Table \ref{tab:finalmodel} represents the enhancements of Flooding(Fld.), Serialization with value(val.) and GP, respectively. The best result is 73.1\% on Dev. offline.

\begin{table}[h]%
  \centering
  \begin{tabular}{cc}
  \hline
    model & Dev.\\
    \hline
    RAT-SQL+${\bf G{\scriptstyle RAPPA}}$ & $71.5\pm0.2$\\
    RAT-SQL+${\bf G{\scriptstyle RAPPA}}$ with Fld. & $71.4\pm0.7$\\
    RAT-SQL+${\bf G{\scriptstyle RAPPA}}$ with Fld. val.  & $71.8\pm0.6$\\
    RAT-SQL+${\bf G{\scriptstyle RAPPA}}$ with Fld. val. GP & $72.5\pm0.6$\\
    \hline
\end{tabular}
\caption{Our final results. Fld. represents Flooding. val. means serialization with value.}
\label{tab:finalmodel}
\end{table}

\begin{table}[h]%
  \centering
  \begin{tabular}{ccc}
    \hline
    model & Dev. & Test\\
    \hline
    RAT-SQL+${\bf G{\scriptstyle RAPPA}}$ \cite{yu2020GRAPPA} & 73.4 & 69.6\\
    RAT-SQL+${\bf G{\scriptstyle RAPPA}}$+GP (Ours) & \bf{72.8} & \bf{69.8}\\
    \hline
\end{tabular}
\caption{ Comparison of the results between RAT-SQL+${\bf G{\scriptstyle RAPPA}}$ and RAT-SQL+${\bf G{\scriptstyle RAPPA}}$+GP}
\label{tab:ratcom}
\end{table}

The ultimate result on Spider is 72.8\% on Dev. and 69.8\% on Test. Compared to the result of RAT-SQL+${\bf G{\scriptstyle RAPPA}}$, the Dev. and Test. The results of RAT-SQL+${\bf G{\scriptstyle RAPPA}}$+GP are much closer indicating that our model is more robust, as shown in Table \ref{tab:ratcom}.

\section{Conclusion}

Since most researches concentrate on natural language generation in Text-to-SQL tasks, SNM was utilized here to analyze the target programming language’s syntax. To reduce SQL grammar errors in the decoder process, we proposed a new framework called GP, for pre-training parameters on the decoder side. Questions are appended by values when they match the word exactly. Schema information is enriched as the input of encoding By averaging the embeddings of all tokens' vector from the column or table instead of the first and last token. Ultimately, we adopted flooding level to avoid local minimum loss in the training procedure. The results proved that this method possesses a greater performance on the Spider dataset. It is also beneficial for other context-free grammar representation tasks. Furthermore, since parameter tuning is a complex task, a tiny difference of parameters, especially learning rate, can result in completely different results. This model still has a high probability for further improvement, thus, some tuning methods will be assessed in the future.

\acks{The authors thank Manfang Wu, Jian Cai for their assistance and advice.  We also acknowledge Xuefeng Li and our anonymous reviewers for their comments.  The Big Data Lab is operated by OneConnect Financial Technology.
}



\vskip 0.2in
\bibliography{main}
\bibliographystyle{theapa}

\end{document}